\documentclass{article}
\usepackage{ijcai26}

\usepackage{times}
\usepackage{soul}
\usepackage{url}
\usepackage[hidelinks]{hyperref}
\usepackage[utf8]{inputenc}
\usepackage[small]{caption}
\usepackage{graphicx}
\usepackage{amsmath}
\usepackage{amsthm}
\usepackage{booktabs}
\usepackage{algorithm}
\usepackage{algorithmic}
\usepackage[switch]{lineno}
\usepackage{multirow}
\usepackage{tabularx}

\title{ShuttleEnv: An Interactive Data-Driven RL Environment for Badminton Strategy Modeling}

\author{
Ang Li$^1$ \and
Xinyang Gong$^1$ \and
Bozhou Chen$^1$ \and
Yunlong Lu$^1$ \and
Jiaming Ji$^2$\and
Yongyi Wang$^1$ \and
Yaodong Yang$^2$\textsuperscript{*}\and
Wenxin Li$^1$\textsuperscript{*}\\
\footnotesize{* Corresponding author}
\affiliations
$^1$School of Computer Science, Peking University\and
$^2$Institute for Artificial Intelligence, Peking University\\
\emails
1901111291@pku.edu.cn, bre@stu.pku.edu.cn, 2301111899@stu.pku.edu.cn, luyunlong@pku.edu.cn, jiamg.ji@stu.pku.edu.cn, wangyongyi@pku.edu.cn, yaodong.yang@pku.edu.cn, lwx@pku.edu.cn
}

\begin{document}

\maketitle

\begin{abstract}
We present \textbf{ShuttleEnv}, an interactive and data-driven simulation environment for badminton, designed to support reinforcement learning and strategic behavior analysis in fast-paced adversarial sports. The environment is grounded in elite-player match data and employs explicit probabilistic models to simulate rally-level dynamics, enabling realistic and interpretable agent-opponent interactions without relying on physics-based simulation. In this demonstration, we showcase multiple trained agents within ShuttleEnv and provide live, step-by-step visualization of badminton rallies, allowing attendees to explore different play styles, observe emergent strategies, and interactively analyze decision-making behaviors. ShuttleEnv serves as a reusable platform for research, visualization, and demonstration of intelligent agents in sports AI. Our \textbf{ShuttleEnv} demo video \textbf{URL}:  \url{https://drive.google.com/file/d/1hTR4P16U27H2O0-w316bR73pxE2ucczX/view}
\end{abstract}

\section{Introduction}

Badminton is a fast-paced, adversarial one-on-one sport that requires players to make rapid tactical decisions under partial information and strong opponent pressure. Effective play depends not only on physical execution, but also on anticipating an opponent’s intent, selecting appropriate shot types, and positioning strategically within extremely short time horizons. These characteristics make badminton a compelling yet underexplored testbed for artificial intelligence research on sequential decision-making and opponent modeling.

Existing computational studies on badminton have largely focused on supervised learning tasks, such as stroke forecasting and movement prediction from historical match data~\cite{wang2022shuttlenet,chang2023will,wang2023benchmarking,ibh2024stroke}. While these approaches provide valuable descriptive insights into player behavior, they do not support interactive decision-making or online policy improvement. More recent efforts have introduced reinforcement learning environments for sports, most notably in team-based settings such as football~\cite{kurach2019google,liu2021towards} and recent badminton-specific platforms~\cite{wang2024coachai}. However, team sports primarily emphasize coordination among multiple agents, whereas badminton highlights high-frequency, fine-grained tactical interactions between two opponents. Physics-based simulators, on the other hand, often incur substantial modeling complexity and sim-to-real challenges, limiting their accessibility and flexibility for rapid experimentation.

As a result, there remains a lack of interactive, data-driven reinforcement learning environments that enable researchers to train, visualize, and analyze strategic behaviors in badminton. In particular, such an environment should allow agents to engage in realistic rallies, expose interpretable decision dynamics, and support human-in-the-loop exploration, rather than serving solely as a benchmarking platform.

In this demonstration, we present ShuttleEnv, an interactive reinforcement learning environment for badminton grounded in elite-player match data and explicit probabilistic rally dynamics.
Our contributions are threefold:
\begin{itemize}
    \item An interactive environment supporting reinforcement learning, enabling agents to engage in rally-level decision-making with interpretable, data-driven transition dynamics.
    \item A manually collected and annotated fine-grained badminton dataset, from which we derive both imitation learning policies and two learned transition models that define environment dynamics.
    \item Extensive RL agent implementations and integrated visualization tools, allowing qualitative analysis of learned strategies and interactive exploration of agent behaviors.
\end{itemize}
Together, these components establish ShuttleEnv as a reusable and demonstrable platform for studying strategic decision-making in adversarial sports.

\section{System Overview of ShuttleEnv}
\begin{figure}[ht]
    \centering
    \includegraphics[width=\linewidth]{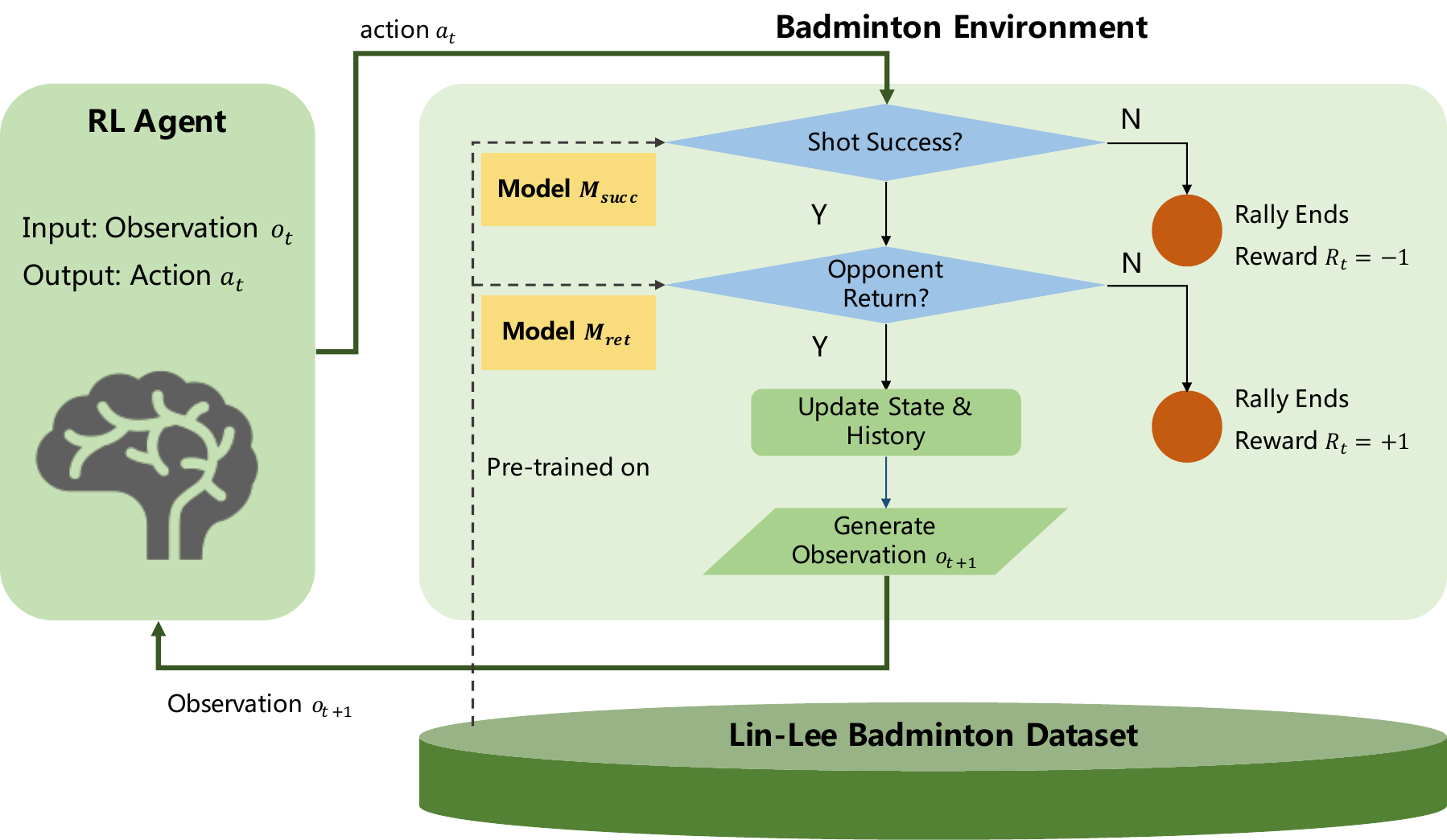} 
    \caption{The workflow of our data-driven badminton environment. The RL agent interacts with the environment in a standard loop. Inside the environment, the outcome of an action $a_t$ is determined by a two-stage probabilistic process, driven by two pre-trained models ($M_{\text{succ}}$ and $M_{\text{ret}}$) which leverage the elite-level badminton match data. The agent receives a new observation $o_{t+1}$ generated from the action history window to make the next decision and a sparse reward $R_t$ at the end of a rally.}
    \label{fig:env_architecture}
\end{figure}
As shown in Figure~\ref{fig:env_architecture}, ShuttleEnv is designed as an interactive reinforcement learning environment that models badminton rallies at the tactical decision level. The environment follows a standard agent-environment interaction loop while abstracting away low-level physics in favor of data-driven, interpretable rally dynamics.

At each decision step, the agent selects a badminton action that represents a high-level tactical choice. An action consists of multiple components, including shot type (e.g., smash, drop, clear, drive, net shot, lift), target court area, shot height, and optional execution attributes such as backhand or around-the-head strokes. These components jointly define the intent of a shot rather than its precise physical trajectory, enabling the environment to focus on strategic decision-making under uncertainty.

Once an action is executed, ShuttleEnv determines the outcome of the rally through a two-stage probabilistic process learned from elite-player match data. First, a shot success model evaluates whether the selected action results in a valid shot or immediately ends the rally due to errors such as hitting the net or out of bounds. If the shot succeeds, a defense outcome model then determines whether the opponent successfully returns the shot or fails to do so, resulting in a point-ending outcome.

The reward signal in ShuttleEnv is sparse and defined at the rally level. The agent receives a positive reward when its action ultimately wins the point and a negative reward when it loses the rally, while intermediate steps yield zero reward. This design mirrors real badminton scoring rules and encourages agents to reason over long-term tactical consequences rather than short-term heuristics.

\section{System Details}
\subsection{Lin-Lee Badminton Dataset}
ShuttleEnv is grounded in a manually collected and annotated dataset derived from broadcast recordings of elite-level badminton matches between Lin Dan and Lee Chong Wei. The dataset consists of rally-level sequences segmented at the shot level, covering thousands of annotated actions across multiple matches.

Each shot is labeled with structured tactical attributes, including shot type (e.g., smash, drop, clear, drive, net shot, lift), target court area, shot height, player position, and rally outcome. These annotations enable both supervised action prediction tasks and the estimation of probabilistic transition models used in environment construction.

Unlike publicly released stroke-forecasting benchmarks designed primarily for predictive evaluation, our dataset is collected and curated specifically to support interactive simulation and reinforcement learning. This alignment between data collection and environment design ensures consistency between observed match statistics and simulated rally dynamics.

\subsection{Data-Driven Environment Dynamics}
The environment dynamics in ShuttleEnv are defined through two learned transition models derived from the dataset.

The first model, denoted as $M_{succ}$, estimates the probability that a selected shot is successfully executed without resulting in an immediate error (e.g., hitting the net or landing out of bounds). This model captures execution reliability under different tactical contexts.

The second model, denoted as $M_{ret}$, models the opponent’s ability to return a valid shot. Given the current rally context and executed action, $M_{ret}$ estimates the likelihood that the opponent successfully responds, thereby continuing the rally.

These two models together determine rally evolution in a modular and interpretable manner. Instead of relying on full physics-based simulation, ShuttleEnv resolves point outcomes through sequential evaluation of shot success and defensive return, allowing realistic yet computationally efficient tactical interactions.

Beyond defining the environment transitions, it is important to verify that the learned models capture meaningful tactical regularities from real matches. To validate that the dataset and learned transition mechanisms capture meaningful tactical patterns, we conduct a sequential action prediction task. 
Given recent rally context, including previous states and actions, the model predicts the next stroke type and landing zone.
\begin{table*}[ht]
\centering
\caption{Top-$k$ accuracy for next-action prediction.}
\begin{tabularx}{\linewidth}{l *{4}{>{\centering\arraybackslash}X} *{4}{>{\centering\arraybackslash}X}}
\toprule
& \multicolumn{4}{c}{\textbf{Stroke Type}} 
& \multicolumn{4}{c}{\textbf{Landing Zone}} \\
\cmidrule(lr){2-5} \cmidrule(lr){6-9}
\textbf{Player} & Top-1 & Top-2 & Top-3 & Top-4 
& Top-1 & Top-2 & Top-3 & Top-4 \\
\midrule
Lin Dan 
& 0.473 & 0.664 & 0.785 & 0.880 
& 0.580 & 0.763 & 0.854 & 0.898 \\

Lee Chong Wei 
& 0.438 & 0.689 & 0.816 & 0.871 
& 0.594 & 0.796 & 0.884 & 0.924 \\
\bottomrule
\end{tabularx}
\label{tab:prediction_acc}
\end{table*}
Table~\ref{tab:prediction_acc} reports Top-$k$ prediction accuracy for both stroke type and landing zone prediction.
The increasing performance with larger $k$ indicates structured tactical patterns in the dataset and supports the validity of the learned transition dynamics.

\subsection{RL Agents and Training}
To demonstrate the usability of ShuttleEnv for reinforcement learning research, we implement multiple policy learning approaches within the environment.

First, we construct imitation learning policies using behavioral cloning (BC) trained on the annotated dataset using supervised action prediction. These policies serve as behavior baselines and reflect tactical tendencies observed in elite matches.

Second, we implement reinforcement learning agents using standard policy gradient algorithms, including A2C~\cite{mnih2016asynchronous}, PPO~\cite{schulman2017proximal} and SAC~\cite{haarnoja2018soft}. These agents are trained directly within ShuttleEnv under rally-level sparse rewards, enabling them to discover strategic behaviors through interaction rather than imitation.

Over 1,000 simulated matches against a fixed opponent, BC achieves a win rate of $33.2\%\pm1.2\%$, while A2C reaches $65.8\%\pm2.0\%$. 
PPO achieves $98.3\%\pm2.5\%$, and SAC achieves $90.5\%\pm7.7\%$. The substantial performance gap between imitation and reinforcement learning agents indicates that ShuttleEnv provides a learnable and strategically meaningful training environment. These learned policies are further analyzed through the integrated visualization interface described in Section~\ref{sec: demo}.

\section{Demonstration Scenarios and Visualization}
\label{sec: demo}
ShuttleEnv is designed not only as a reinforcement learning environment, but also as a live interactive platform for exploring badminton strategy and tactical decision-making.
During the demonstration, attendees can observe trained agents exhibiting diverse play styles, including aggressive attack-oriented policies, defensive rally construction, and risk-aware shot selection strategies. By pairing different agents or switching policies in real time, users can directly inspect how strategic differences influence rally tempo, shot distribution, and point outcomes.

\begin{figure}[ht]
    \centering
    \includegraphics[width=\columnwidth]{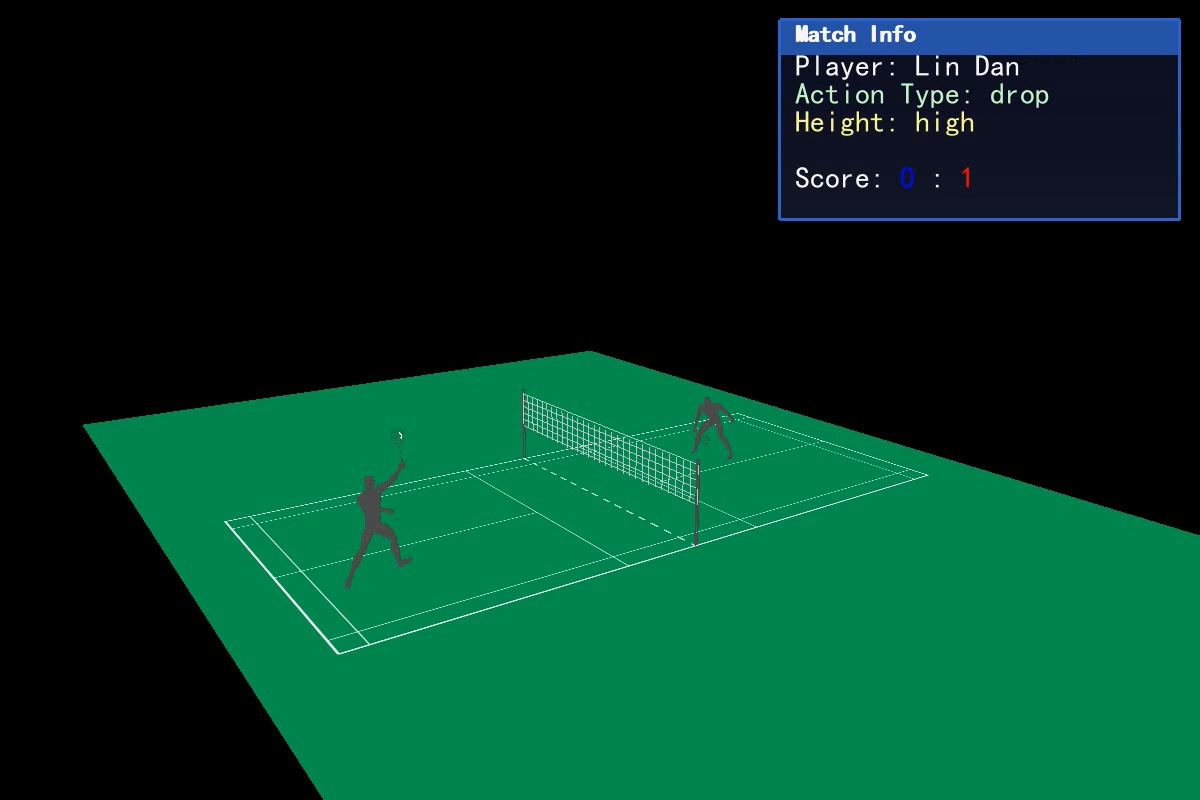}
    \caption{Representative frame from the visualization demo. The rendering overlays rally context (active player, action metadata, and score) on a 3D court scene. The visualization highlights how different shot selections and target areas lead to distinct rally trajectories and outcomes, supporting qualitative inspection of tactical decision-making.}
    \label{fig:visualization_demo}
\end{figure}
A central feature of the system is its fully integrated 3D visualization module (Figure~\ref{fig:visualization_demo}). The environment renders complete rally simulations using articulated humanoid player models equipped with badminton-specific motion primitives. Professional mesh models and stroke animations are combined to produce coherent representations of body orientation and hitting actions. Rather than displaying abstract state transitions alone, ShuttleEnv visually maps tactical decisions—such as shot type and target selection—onto physically interpretable player movements and shuttle trajectories on a realistic court scene.

Contextual overlays further augment the visualization by displaying the active player, selected action components, target landing areas, and current score. This integration of tactical metadata with embodied motion provides a clear bridge between high-level decision policies and observable in-game behavior. As a result, attendees can qualitatively analyze how learned strategies manifest as plausible and interpretable rally dynamics.

The demo interface supports interactive playback control, including pausing, replaying rallies, and dynamically switching between agent configurations. These capabilities allow users to experiment with policy variations and immediately observe their impact, making ShuttleEnv suitable for live demonstration, qualitative analysis, and educational exploration of strategic behavior in adversarial sports.

\section{Discussion and Conclusion}

In this demonstration, we presented ShuttleEnv, an interactive and data-driven reinforcement learning environment for badminton. Grounded in a manually collected elite-match dataset and equipped with learned probabilistic transition models, ShuttleEnv supports rally-level tactical reasoning within a realistic yet computationally efficient framework. By integrating reinforcement learning agents with a 3D visualization system, the platform connects high-level decision policies to interpretable on-court behavior, enabling qualitative analysis and live demonstration of strategic play.

ShuttleEnv adopts an abstraction-based modeling approach rather than full physics simulation. While this design improves efficiency and interpretability, it does not explicitly model detailed biomechanics or shuttle aerodynamics. Moreover, the current dataset focuses on matches between two elite players, which may limit stylistic diversity in simulated scenarios. Future work may expand the dataset, incorporate richer motion representations, and explore multi-agent training settings to further enhance realism and flexibility.

We envision ShuttleEnv as a step toward more transparent and interactive environments for studying strategic intelligence in individual sports. By combining data-driven modeling, reinforcement learning, and embodied 3D visualization, the platform offers a foundation for future research at the intersection of sports analytics, simulation, and artificial intelligence.

\newpage
\bibliographystyle{named}
\bibliography{ijcai26}

\end{document}